\documentclass[10pt,twocolumn,letterpaper]{article}

\usepackage[pagenumbers]{cvpr} 

\usepackage[dvipsnames]{xcolor}

\definecolor{cvprblue}{rgb}{0.21,0.49,0.74}
\usepackage[pagebackref,breaklinks,colorlinks,citecolor=cvprblue]
{hyperref}
\usepackage{graphicx}
\usepackage{amsmath}
\usepackage{amssymb}
\usepackage{booktabs}
\usepackage{textcomp}
\usepackage{multirow}
\usepackage{graphicx}
\usepackage{float}
\usepackage[linesnumbered,ruled,vlined]{algorithm2e}
\usepackage{enumitem}
\usepackage{appendix}

\def\paperID{10697} 
\def\confName{CVPR}
\def\confYear{2024}

\title{Breaking the Black-Box: Confidence-Guided Model Inversion Attack for Distribution Shift}

\author{
    Xinhao Liu, Yingzhao Jiang, Zetao Lin  \\
    \texttt{\small 2210273140, 2210273146, 2310273128@email.szu.edu.cn}
}

\begin{document}
\maketitle
\begin{abstract}
Model inversion attacks (MIAs) seek to infer the private training data of a target classifier by generating synthetic images that reflect the characteristics of the target class through querying the model. However, prior studies have relied on full access to the target model, which is not practical in real-world scenarios. Additionally, existing black-box MIAs assume that the image prior and target model follow the same distribution. However, when confronted with diverse data distribution settings, these methods may result in suboptimal performance in conducting attacks. To address these limitations, this paper proposes a \textbf{C}onfidence-\textbf{G}uided \textbf{M}odel \textbf{I}nversion attack method called CG-MI, which utilizes the latent space of a pre-trained publicly available generative adversarial network (GAN) as prior information and gradient-free optimizer, enabling high-resolution MIAs across different data distributions in a black-box setting. Our experiments demonstrate that our method significantly \textbf{outperforms the SOTA black-box MIA by more than 49\% for Celeba and 58\% for Facescrub in different distribution settings}. Furthermore, our method exhibits the ability to generate high-quality images \textbf{comparable to those produced by white-box attacks}. Our method provides a practical and effective solution for black-box model inversion attacks.
\end{abstract}    
\section{Introduction}
\label{sec:intro}

Privacy protection and attacks have been extensively studied, attracting significant attention within the scientific community\cite{liu2020privacy,wernke2014classification,rigaki2020survey,PrivacySurvey}. Model inversion attacks (MIAs) represent a class of attacks aimed at compromising the privacy protection of models\cite{song2022survey}. MIAs target the retrieval of sensitive information about the model’s training data by leveraging known model outputs, thus putting user privacy at risk. For instance, an attacker may query the output of a facial recognition model and, upon successful exploitation, generate synthetic images that reflect the user’s facial features, thereby violating user privacy.

\begin{figure}[t]
  \centering
   \includegraphics[width=1.0\linewidth]{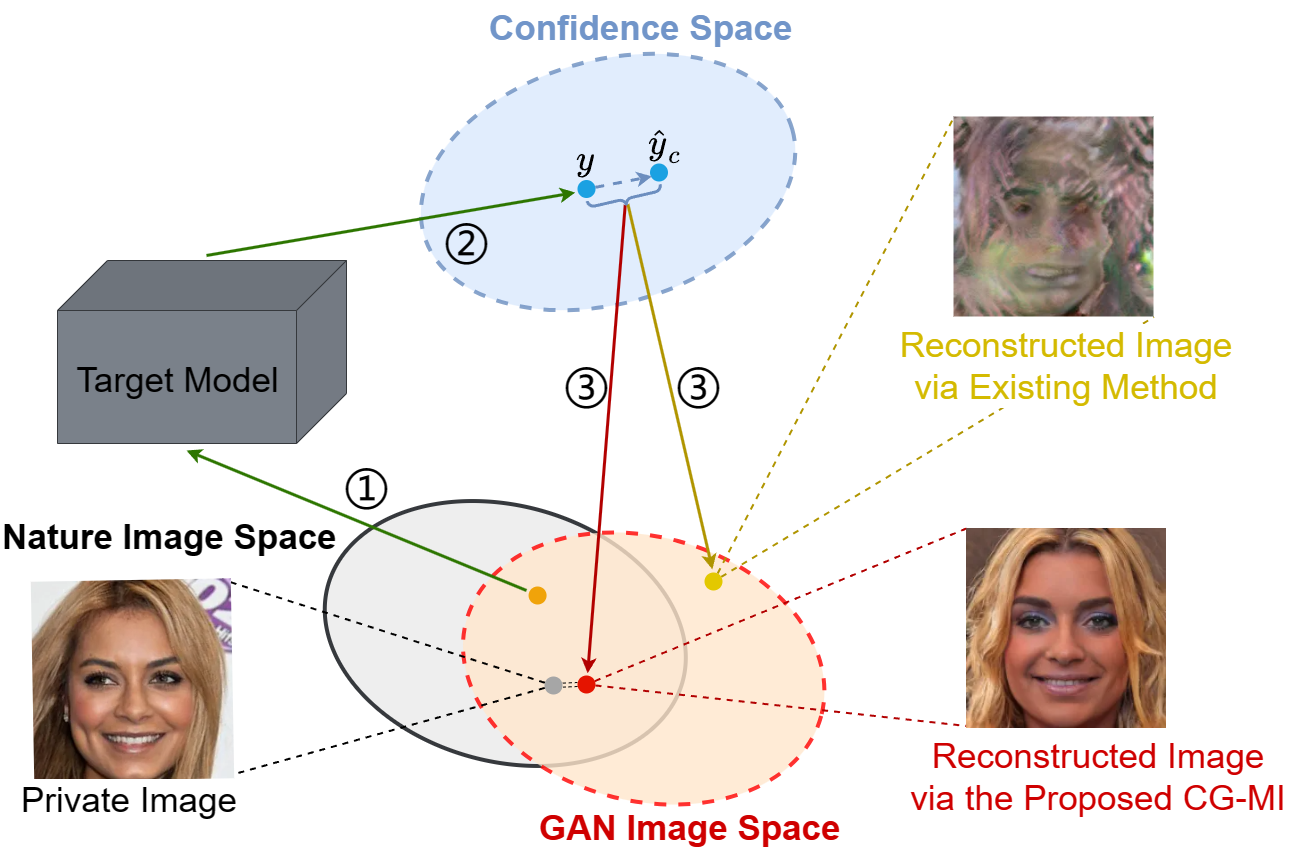}

   \caption{Illustration of private training data leakage for a specific target class \(c\) via the target model output confidence: 
\textcircled{1} Adversary inputs the initially generated image into the target model. 
\textcircled{2} Adversary obtains the model output confidence \(y\). 
\textcircled{3} Adversary attempts to reconstruct the private training data from the confidence \(y\) to \(\hat{y}_c\) in a black-box setting, where \(\hat{y}_c\) is the one-hot vector for class \(c\). The existing method is the optimization result in $W$ space.} 
   \label{abstract_figure}
\end{figure}

Model inversion attacks can currently be categorized into three types: white-box attacks\cite{zhang2020secret,yuan2023pseudo,struppek2022ppa,chen2021knowledge,rethink,vmi}, black-box attacks\cite{han2023reinforcement,black-box,black-box-,black-boxlbmi,mirror}, and label-only attacks\cite{labelonly}, based on the level of access the attacker has to the target model. In the context of white-box attacks, the attacker has complete access to the target model, including its knowledge such as weights and output confidence scores. In the case of black-box attacks, the attacker has limited access and can only utilize the confidence scores provided by the target model without any internal knowledge. In the label-only setting, the attacker can only use the output labels provided by the model.

Currently, there has been significant research focus on white-box MIAs \cite{zhang2020secret,chen2021knowledge}. These attack methods require complete access to the target model for performing MIAs. Moreover, these methods assume that the private training data of the target model and the public data used to train the generative model follow the same distribution, which is not practical in real-world scenarios. To address this challenge, Plug \& Play Attack(PPA) \cite{struppek2022ppa} proposed an independent white-box MIA that works under different data distribution settings. However, it still relies on complete access to the target model. In the black-box domain, Reinforcement Learning-Based Model Inversion attack(RLB-MI) \cite{han2023reinforcement} focused on black-box MIA within the same data distribution using reinforcement learning techniques. However, their attack performance on different data distributions is not satisfactory, and the synthesized images have lower resolution. \textbf{Therefore, a crucial challenge in the black-box MIAs scenario is how to generate high-resolution and effective synthetic images solely based on the confidence scores provided by the target model across different data distributions.}

The challenges in this task can be summarized as follows: Firstly, generating high-resolution synthetic images in a high-dimensional latent space poses optimization difficulties. Secondly, the absence of gradient information in black-box model inversion attacks may lead optimization algorithms to explore GANs' latent space excessively, resulting in the generation of images without meaningful features and leading to the failure of the attack. These challenges hinder the direct application of existing white-box attack method\cite{struppek2022ppa} to black-box scenarios or limit the attack effectiveness in different data distribution scenarios\cite{han2023reinforcement}. The adversary's attack process in a black-box scenario is illustrated in \Cref{abstract_figure}.

To address the limitations described above, our paper proposes CG-MI, a novel approach that achieves MIAs in black-box settings with different data distributions. The main idea is to leverage a pre-trained, target-independent generative adversarial network (GAN)\cite{Gan,stylegan2-ada} as image prior, then employ gradient-free optimization methods to minimize the confidence loss, which measures the matching between the GAN image manifold and the target model. To overcome dimensionality issues and avoid generating meaningless images during the optimization process, we propose a novel objective optimization function. The core idea is to incorporate the mapping network of StyleGAN2\cite{stylegan2-ada} into the gradient-free optimization process, thereby ensuring that the solution of the optimization problem remains within a meaningful exploration space. Extensive experiments demonstrate that our method significantly outperforms existing black-box MIAs and its ability to generate high-quality images comparable to white-box attacks. Our main contributions are as follows:
\begin{enumerate}[label=\textbullet]
    \item We present a novel approach to black-box MIAs by utilizing gradient-free optimizer-based method. Our method enables MIAs in black-box scenarios, accommodating various data distributions and generate high-resolution synthesis images.
    \item We propose the concept of \textit{synthesis image transferability in model inversion}, analysis its impact on MIAs, and address this issue by designing a novel objective optimization function.
    \item We demonstrate on different datasets and models with the proposed CG-MI. Compared to state-of-the-art black-box attack methods, our approach significantly improves attack performance. Furthermore, visual comparisons indicate comparable synthesis image quality to white-box approaches.
\end{enumerate}

Our work shows that in more challenging scenarios, MIAs still can lead to the leakage of private information from DNNs. 

\begin{figure*}[!ht]
  \centering
   \includegraphics[width=1.0\linewidth]{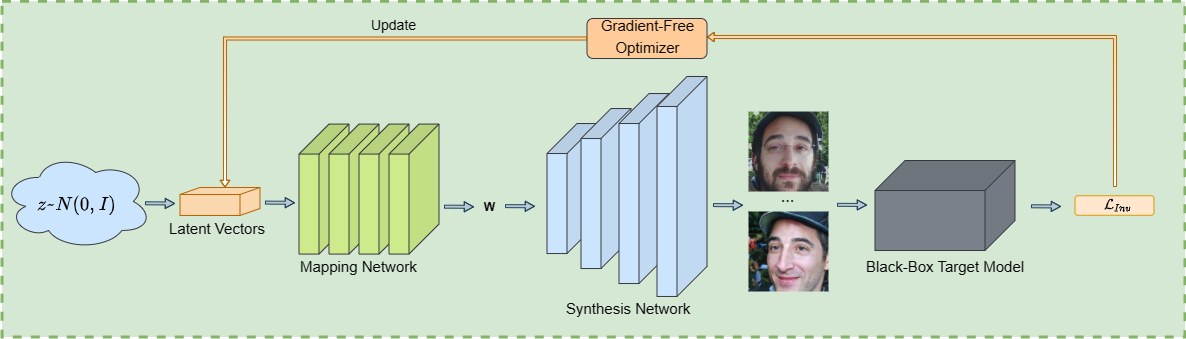}

   \caption{The overview of the proposed attack.
Latent vectors z are sampled from a standard normal distribution $N(0,1)$. These latent vectors are then passed through a mapping network to obtain style vectors w. The style vectors w are subsequently fed into a synthesis network to generate corresponding images. These generated images are further inputted into the target model, and the loss is calculated based on the objective function. The latent vectors z are updated using a gradient-free optimization method. This process continues until we obtain optimized synthesized images.}
   \label{fig:overview}
\end{figure*}
\section{Related Work}
MIAs can be viewed as an optimization problem, where the objective is to maximize the confidence scores of a given target class in order to generate images that reveal sensitive data features. MIAs were first introduced by \cite{firstpaper2014} for attacking linear regression models. Subsequently, \cite{first-paper} proposed a gradient descent-based algorithm to attack shallow networks. In the following sections, we will introduce recent attack methods based on the types of MIAs.

 \textbf{White-Box MIAs.}
 White-box MIAs leverage full access to the target model and utilize gradient-based optimization techniques. \cite{zhang2020secret} was the first to propose a generative attack method for MIAs, to enable MIAs for deeper networks. They trained a DCGAN \cite{DCGAN} on publicly available data that had no overlap with the private training data, and conducted the attack by optimizing the latent input vector $z$ of the GAN to attack the target model trained on the private data. Subsequent work, such as \cite{chen2021knowledge}, incorporated soft labels of the target model to guide the GAN training and allow the generator to learn the latent distribution, enabling specific GANs for MIAs. Furthermore, \cite{yuan2023pseudo} introduced a Conditional GAN\cite{cgan} as an image prior model for MIAs, addressing the issue of the generator in \cite{chen2021knowledge} not fully utilizing the target model’s knowledge. Additionally, \cite{rethink} proposed a method that directly maximizes confidence scores instead of minimizing negative log-likelihood scores, aiming to improve the attack performance of \cite{zhang2020secret} and \cite{chen2021knowledge}. \cite{vmi} introduced a variation-based MIAs using StyleGAN2 \cite{stylegan2}, capable of generating high-resolution images that reflect the target model’s private training data. Moreover, \cite{struppek2022ppa} introduced a dataset-agnostic MIAs approach utilizing pre-trained StyleGAN2 \cite{stylegan2-ada} models. This method targets the vulnerability of prior works to dataset distribution shifts, aiming to address this concern.

\textbf{Black-Box MIAs.} 
Black-box MIAs require access to the confidence scores of the target model. In black-box MIAs, \cite{black-box-} proposed a method that simultaneously trains a GAN and a surrogate model. The GAN is used to generate inputs similar to the private training data, while the surrogate model imitates the behavior of the target model for inversion attacks. Additionally, \cite{black-box} attempted to recover faces from deep feature vectors of a face recognition model in a black-box setting without prior knowledge. Another attack model was introduced by \cite{black-boxlbmi}, where they perform MIAs by swapping the input and prediction vectors of the target model. Furthermore, \cite{mirror} proposed a black-box MIAs method based on StyleGAN, using a classical genetic algorithm for optimization. Recently, \cite{han2023reinforcement} presented a reinforcement learning-based approach for MIAs, where the confidence scores of the target model’s outputs serve as rewards.

\textbf{Label-Only MIAs.}
Label-only MIAs focus on querying the model to obtain hard labels without confidence scores. \cite{labelonly} introduces an algorithm called Boundary Repulsion Model Inversion (Brep-MI). The core idea of this algorithm is to evaluate the model’s predicted labels on a spherical surface and then estimate the direction towards the center of the target class to generate the most representative image.
\section{Threat Model}

\textbf{Attack goal.} 
When the target model $M$ is a facial recognition classifier, the aim of MIAs is to exploit the attacker’s access to generate facial images that reflect the features of a specific class, represented as $c \in C$, $C$ represents all classes. 

\textbf{Model Knowledge.} 
In white-box MIAs, the attacker possesses the ability to download the model and exploit its weights and confidence information for launching the attack. In contrast, black-box MIAs restrict the attacker to using only the confidence scores provided by the target model. In label-only MIAs, the attacker is limited to utilizing the model’s output labels. Our research focus on the black-box MIAs.

\textbf{Data Knowledge.} 
In the majority of existing white-box MIAs\cite{chen2021knowledge,zhang2020secret,yuan2023pseudo} and black-box MIAs \cite{han2023reinforcement}, they assume that the attacker can launch attacks on data from the same distribution, i.e., $P(X_{prior}) = P(X_{target})$, $P(X_{prior})$ represents the distribution of image priors and $P(X_{target})$ represents the distribution of target model. In our work, we relax the assumption that the attacker is only aware of the targeted model’s classification task, such as facial recognition, under the setting where $P(X_{prior}) \neq P(X_{target})$.
\section{Methodology}

\subsection{Background}

\textbf{Problem Formulation.} 
We define the target classification model as $M$, with $x$ representing the input image to the target model and $c$ denoting the target class for the attack. To obtain a synthesized image $x^\ast$ that reflects the private features of the target class $c$, we optimize the following loss function:

\begin{equation}
  x^\ast=arg{\min_x}{\mathcal{L}
  \left(M\left(x\right),c\right)}
  \label{eq:problem formula}
\end{equation}

Here, $\mathcal{L}$ can be the cross-entropy loss or other suitable loss functions. The purpose of this loss is to directly optimize the image $x$ in order to leak the private training data of the model. As directly optimizing the high-dimensional vector $x$ is not efficient. The following section will delve into generative MIAs.

\textbf{Generative Model Inversion Attacks.}
The idea of training a generative model as an image prior to optimize the latent vector $z$ in the GAN for image synthesis was first introduced by \cite{zhang2020secret}. This approach addresses the problem discussed in \cite{first-paper}, where directly optimizing $x$ in the high-dimensional, nonlinear, and non-convex solution space when attacking deep neural networks can lead to the generation of meaningless results. Their method involves training a DCGAN \cite{DCGAN} on publicly available data that does not overlap with the private training data, and then optimizing the latent input vector of the GAN to attack a target model trained on private data. By introducing the image prior GAN, the optimization problem in \Cref{eq:problem formula} can be expressed as:

\begin{equation}
z^\ast=arg\min_z\mathcal{L}\left(M\left(G\left(z\right)\right),c\right)
\label{generativemi}
\end{equation}

After optimizing \Cref{generativemi} to obtain $z^\ast$, then input $z^\ast$ into the GAN\cite{DCGAN} to generate the synthesized image $x^\ast = G(z^\ast)$. This approach helps mitigate the issue of generating meaningless images to a certain extent.

\subsection{Breaking the Black-Box}

In this section, we will present our approach, Confidence-Guided Model Inversion (CG-MI), for attacking different data distribution models in a black-box scenario. An overview of CG-MI is illustrated in \Cref{fig:overview}.

\textbf{Pre-Trained Publicly Available Image Prior.}
In the architecture of a GAN\cite{DCGAN,biggan,stylegan2-ada}, the generator model learns to map latent vectors sampled from a simple distribution (e.g., Gaussian or uniform distribution) denoted as $z$, to the generated image $x$. However, StyleGAN2\cite{stylegan2-ada} consists of two main components: $G_{mapping}$ and $G_{synthesis}$. In StyleGAN2, the latent vector $z$ is first transformed into a style vector w using a non-linear mapping network $f$, implemented as an $8$-layer Multi-Layer Perceptron (MLP). The style vector $w$ is then transformed into a synthesized image $x$. Specifically, a mapping network $G_{mapping}: Z \to W$, where $z \in Z$ and $z\sim \mathcal{N}(0,1)$, a synthesis network $G_{synthesis}: W \to X$, generating the corresponding image $x$ based on the input style vector $w \in W$. Previous works on PPA\cite{struppek2022ppa} have demonstrated the tremendous potential of StyleGAN2 across diverse data distributions. In our paper, while ensuring the independence between the generator model and the target model, we leverage a pre-trained publicly available StyleGAN2 model as our image prior to perform attacks between different data distributions.

\textbf{Synthesis Image Transferability in MIAs.}
Consider two well-trained face recognition models, denoted as $M_1$ and $M_2$, trained on the same dataset $P(X_{target})$. Let $x^*$ be a synthetic image generated through an attack on $M_1$, classification result as $max(M_1(x^*)) = c$. In the ideal scenario of transferability, the generated image $x^*$ satisfies both $max(M_1(x^*)) = max(M_2(x^*)) = c$. This indicates that both $M_1$ and $M_2$ are able to recognize $x^*$ and classify it into the target class $c$. Conversely, if $x^*$ lacks transferability, it may result in $max(M_2(x^*)) \neq c$, leading to attack failure. 
\begin{figure}[t]
  \centering
   \includegraphics[width=1.0\linewidth]{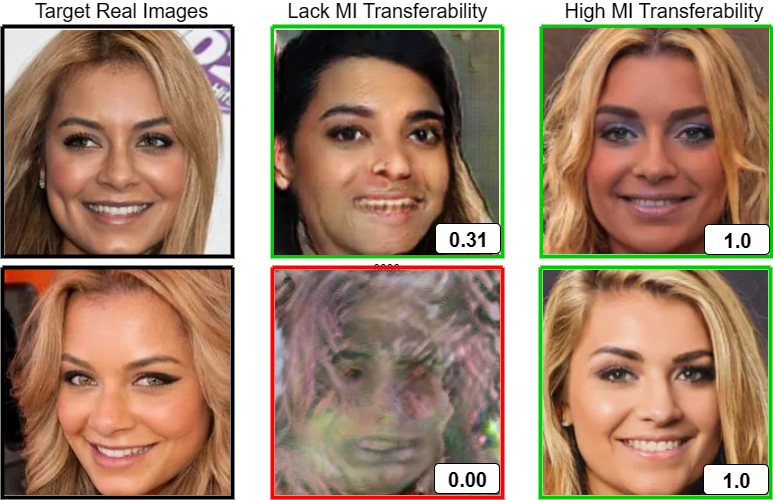}

   \caption{In the comparison between lack synthesis image transferability and high synthesis image transferability in MIAs targeting the same identity. The first image in the second column represents an attack\cite{han2023reinforcement} generated using DCGAN, while the second image in the second column represents an attack result achieved by combining a objective function proposed by PPA\cite{struppek2022ppa} with gradient-free optimization. The third column displays attack results generated by combining our proposed objective function with gradient-free optimization algorithm. The scores inside the pictures represent the confidence scores provided by the evaluation model.}
   \label{fig:loss_comparision}
\end{figure}

\textbf{Enhancing Synthesized Images Transferability with Meaningful Exploration.}
In previous works, such as Brep-MI \cite{labelonly} and RLB-MI \cite{han2023reinforcement}, the chosen image prior model was DCGAN\cite{DCGAN}. However, the overall quality of the synthetic images generated by DCGAN is not satisfactory, as illustrated in \Cref{fig:loss_comparision}, thereby undermining the transferability of the attack. In the white-box PPA\cite{struppek2022ppa} work, StyleGAN2 \cite{stylegan2-ada} was selected as a replacement for DCGAN. They achieved success by optimizing the $w[bs, 14, 512]$ vectors, where $bs$ represents the batch size, by leveraging the target model's weights and utilizing a gradient descent algorithm. The objective function for their method is shown in \Cref{object_function_ppa}.
\begin{equation}
    \begin{aligned}
         w^* = & \arg\min_{w} \mathcal{L}(M(G_{synthesis}(w), c)) 
    \end{aligned}
    \label{object_function_ppa}
\end{equation}
where $G_{synthesis}$ is the synthesis network of StyleGAN2, $c$ is the target class for the attack, $M$ denotes the target model, $w$ is the style vector of the generative model. However, in a black-box scenario, the use of gradient-free optimization algorithms for the direct optimization of the equation above encounters several model inversion issues, leading to the generation of images lacking meaningful features. The main issue arises because gradient-free optimization algorithms focus solely on minimizing the loss function without considering the preservation of the underlying structure of the GAN latent space. Without constraints, they can deviate from the natural image space of the GAN, leading to synthesized images that receive high confidence scores from the target model while the evaluation model assigns them low confidence scores. Moreover, the high dimensionality of $w$ hampers the efficiency of gradient-free optimization algorithms.

To address the issue of generating images lack meaningful features and to reduce the dimensionality of the optimization variables, we focus on the $z$ vectors before they enter the input mapping network. Regardless of how $z$ is changed during the optimization process, the mapping network consistently maps $z$ to a meaningful latent space, thereby avoiding the generation of meaningless images. In comparision to the high dimensionality of $w$, $z$ has dimensions of $[bs, 512]$. Specifically, with a pre-trained StyleGAN2, we aim to solve the following optimization problem:
\begin{equation}
    \begin{aligned}
         z^* = & \arg\min_{z} \mathcal{L}_{Inv}(M(G_{synthesis}(G_{mapping}(z)), c)) 
    \end{aligned}
    \label{object_function}
\end{equation}
where $z \in \mathbb{R}^k$, $\mathcal{L}_{Inv}$ is confidence matching loss, $G_{mapping}$ is the mapping network of stylegan2. By incorporating a mapping network into the optimization process and utilizing a gradient-free optimization algorithm, we have successfully transitioned the problem of solving the black-box MIA from the unrestricted latent space to a space characterized by meaningful facial features. 

\textbf{Confidence Matching Loss.}
The Confidence Matching Loss encourages the solver to find images that can reflect the characteristics of the private training data in the image prior's latent space by minimizing the loss between the target model output confidence $M(x)_c$ and the label $c$. We explore the following confidence matching loss: (1) Cross-Entropy Loss\cite{chen2021knowledge}; (2) Max-Margin Loss\cite{yuan2023pseudo}; and (3) Poincaré Loss\cite{struppek2022ppa}. We performed several comparisons in \Cref{Ablation Study Table} and we final chose poincar\'{e} loss. For specific details of the loss functions, please refer to Appendix \Cref{loss}.

\textbf{Gradient-free Optimizer.}
The optimization problem in \Cref{object_function} is a non-linear and non-convex problem. Choosing a suitable optimization algorithm is crucial for achieving good performance. In this study, we consider the Covariance Matrix Adaptation Evolution Strategy (CMA-ES)\cite{hansen2016cma}, a gradient-free optimization algorithm that is particularly well-suited for high-dimensional problems. CMA-ES is a variant of evolutionary strategies \cite{DE,pycma} and utilizes an adaptive covariance matrix to optimize the probability distribution. We initiate the optimization process by inputting an initial latent vector z. After obtaining the optimized $z^*$, we can generate an image $x^*$ that reflects the private training data features with a target model class label of $c$ using $G_{synthesis}(G_{mapping}(z^*))$.  We maintain the parameter settings as specified in \cite{pycma} for the parameters in the algorithm. Please refer to \Cref{weidaima} for details.

\begin{figure*}[!ht]
  \centering
   \includegraphics[width=1.0\linewidth]{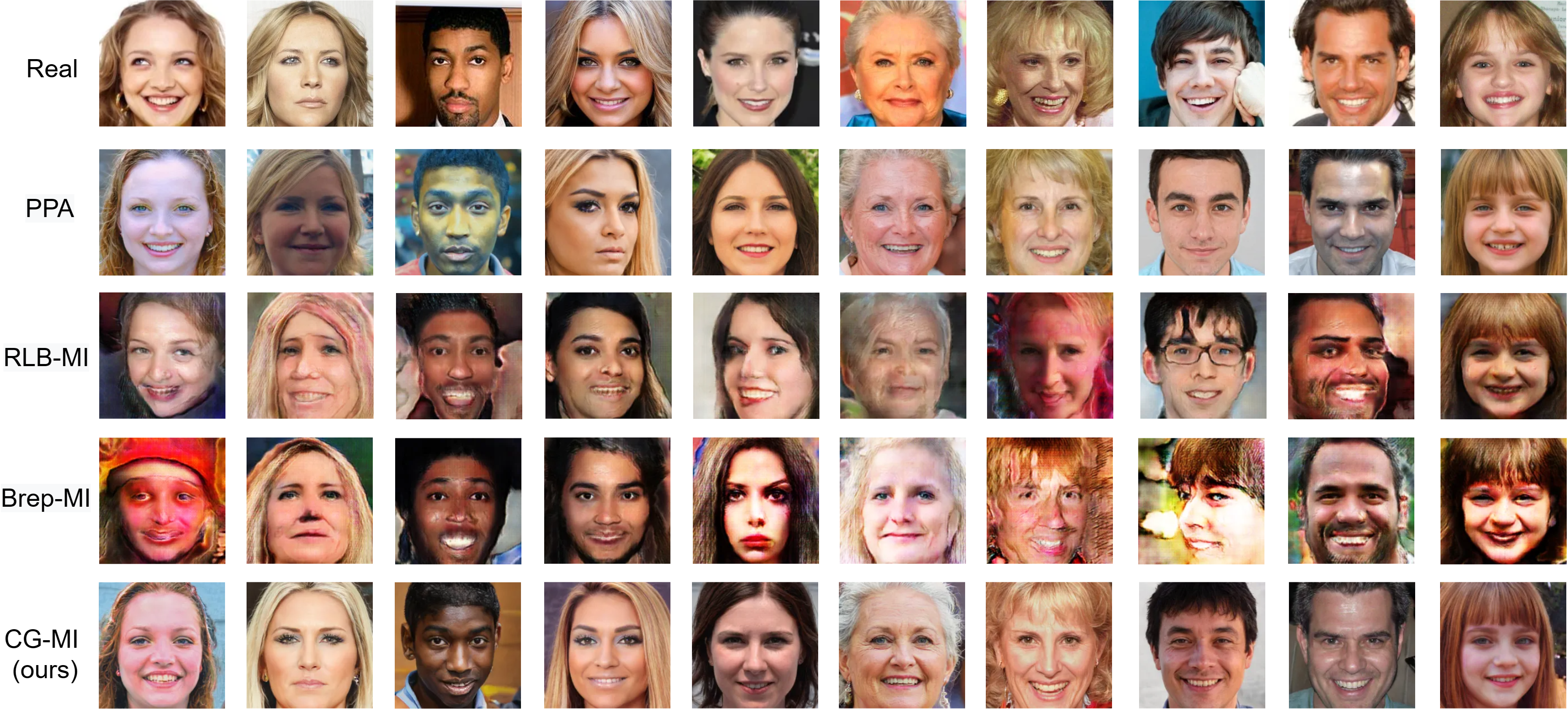}
   \caption{We present a visual comparison of the attack results for different methods in the scenario where the $P(X_{prior})$ = FFHQ, $P(X_{target})$ = CelebA and the target model architecture is Resnet18. The first row shows ground truth images of target class. The second row represents PPA\cite{struppek2022ppa}, the third row represents RLB-MI\cite{han2023reinforcement}, and the fourth row represents Brep-MI\cite{labelonly}. The last row introduces our proposed method, CG-MI.} 
   \label{comparisonresult}
\end{figure*}

\textbf{Transformation-based Selection.}
Following \cite{struppek2022ppa}, we utilize transformation-based selection to choose the images from the optimized population that best reflect the private training data. Let $x = \{x_1, x_2\}$ denote the set of optimized synthetic images. We define a transformation operation $T$, which includes scaling, modifying aspect ratio, and random horizontal flipping. By applying $T$ to $x_1$ and $x_2$, we obtain transformed images, $ T(x_1)$ and $T(x_2)$, which are then input to the target model, yielding new confidence scores, $l_1$ and $l_2$. Typically, if $x_1$ captures the target class features more effectively than $x_2$, the confidence scores after transformations satisfy $l_1 > l_2$. 
\begin{equation}
{E}\left[M\left(T\left(x\right)\right)_c\right]\ \approx\ \frac{1}{N}\ \sum_{i=1}^{N}{M\left(T\left(x\right)\right)_c\ }
    \label{transformation}
\end{equation}
Hence, the image with the highest logit after transformations, denoted by $x_1$, is selected as the final image. In \Cref{transformation}, the Monte Carlo estimation method are employed, where $N$ represents the number of applied transformations.
\section{Experiments}
This section begins with a comprehensive explanation of our experimental setup. Subsequently, we assess the effectiveness of our CG-MI attack by considering different factors such as the performance of various MIA methods in different scenarios, different datasets and different target models.

\subsection{Experimental Settings}

\textbf{Datasets.}
In our experiments, we focus on five datasets: CelebA\cite{celeba}, FFHQ\cite{FFHQ}, FaceScrub\cite{facescrub}, AFHQ Dogs\cite{afhq}, Metfaces\cite{metafaces} and Stanford Dogs\cite{stanforddogs}. For the purpose of conducting attack evaluations, we trained our target models on CelebA, FaceScrub, and Stanford Dogs datasets. As a prior for image synthesis, we utilized pre-trained stylegan2 models on FFHQ, Metfaces and AFHQ Dogs datasets, enabling us to perform attacks on different data distributions. 

\textbf{Models.}
Our models are divided into target models and evaluation models. To facilitate a fair comparison, we conducted attack experiments on several popular network architectures, including Resnet\cite{resnet} and DenseNet\cite{densenet}, while selecting Inception-V3\cite{inceptionv3} as the evaluation model. For the facial recognition task, we trained Resnet18, Resnet152, Densenet169, and InceptionV3 models on the CelebA and FaceScrub datasets, respectively. Similarly, for the dog breed classification task, we trained Resnet18, Resnet152, Densenet169, and InceptionV3 models on the Stanford dogs dataset. To facilitate comparison with prior work, we attacked the Resnet18 model trained on the CelebA and FaceScrub datasets for comparative experiments. The details of the data partition used for training the target model and the parameters for model training, the attack process, and comparative experiments can be referenced in \Cref{attackimplementation}.

\textbf{Evaluation Metrics.}
To align with prior work, we followed PPA\cite{struppek2022ppa} to calculate various evaluation metrics. Firstly, we trained an independent Inception-v3 evaluation model on the training data of the target model. Then, we used the evaluation model to predict labels on the generated attack results and computed the TOP-1 and TOP-5 accuracy for the target class.

Next, we computed the shortest feature distance from each generated image to any training sample in the target class and denoted the average distance as $\delta_{eval}$. The distance was measured using the squared L2 distance between activation layers in the penultimate layer of the evaluation model. For facial images, we utilized a pre-trained FaceNet model \cite{facenet} to measure the feature distance $\delta_{Face}$. Lower values indicate that the attack results are visually closer to the training data.

The third metric is the FID (Fréchet Inception Distance) score \cite{fid}. FID calculates the distance between the feature vectors of the generated attack results and the training data of the target. The feature vectors are extracted using an Inception-v3 model trained on ImageNet \cite{imagenet}. A lower FID score indicates a higher similarity between the two datasets.

\subsection{Experimental Results}
\textbf{Comparision with Previous MIAs Approachs.}
We compared CG-MI with various MIAs methods in different scenarios, including white-box, black-box, and MIAs in the label-only setting. For white-box MIAs, we used PPA\cite{struppek2022ppa} as the baseline method. In contrast to previous white-box attack methods (such as \cite{zhang2020secret,chen2021knowledge,yuan2023pseudo}), PPA focuses on scenarios with different data distributions, image priors, and independence from the target model, making it more meaningful for comparison. For the black-box attack scenario, we selected RLB-MI\cite{han2023reinforcement} and Brep-MI\cite{labelonly} as baseline methods, representing state-of-the-art MIA methods in black-box attacks and the label-only setting, respectively.

To ensure fair comparison, we trained the Resnet18 model on the CelebA and Facescrub datasets for conducting comparative experiments using different MIAs methods. In our CG-MI method, we first ran the CMA-ES optimization algorithm multiple times for each class, generating a batch of 200 synthetic images. This approach ensured the stability and reliability of the results and reduced the impact of randomness. Next, we adopted a transformation selection strategy to choose the most representative 50 images from the optimized batch of 200 synthetic images for evaluation. For the other MIAs methods, we combined the characteristics of each attack method and ran it multiple times to generate a total of 200 synthetic images. Additionally, these methods also incorporate a transformation-based selection strategy to choose 50 images. We then used the same evaluation models and metrics to assess all the methods. 

It is worth mentioning that RLB-MI and Brep-MI both employ the same GAN\cite{DCGAN} structure, which is designed specifically for generating low-resolution 64x64 pixel images. To ensure a fair comparison, we made adjustments to the aforementioned methods by using a deeper GAN architecture capable of generating higher-resolution images\cite{struppek2022ppa}. Firstly, we added two additional upsampling blocks (consisting of a transpose convolution layer and a batch normalization layer) to the generator. We also expanded the discriminator by adding two convolution blocks, with each block consisting of a convolution layer and an instance normalization layer. Subsequently, we trained the modified generator on the FFHQ256 dataset to generate 256x256 pixel images. We then utilized this enhanced GAN as the image prior for RLB-MI and Brep-MI. By adapting the GAN architecture and training on higher-resolution images, we ensured that the comparison between CG-MI and RLB-MI/Brep-MI was carried out on a level playing field.

\begin{table}[t]
\renewcommand{\arraystretch}{1.5}
\resizebox{1.0\columnwidth}{!}{%
\begin{tabular}{@{}cccllccc@{}}
\toprule
\multicolumn{1}{l}{} &
   \textbf{Type} &
   \textbf{Method} &
   \textbf{↑acc@1} &
   \textbf{↑acc@5} &
   ↓$\delta_{\textbf{face}}$ &
   ↓$\delta_{\textbf{$\textbf{eval}$}}$ &
   $\textbf{↓$\textbf{FID}$} $  \\
\midrule
\multirow{4}{*}{\rotatebox{90}{\textbf{CelebA}}}    & White-box                  & PPA     & \textbf{88.28\%}                     & \textbf{97.34\%}                     & 0.6992 & 283.89 & 40.43  \\ \cline{2-8} 
                           & \multirow{2}{*}{Black-box} & RLB-MI  & 29.25\%                     & 53.77\%                     & 1.0740 & 358.18 & 101.86 \\
                           &                            & \textbf{CG-MI(Ours)}   & \textbf{77.86\%}                     & \textbf{94.16\%}                     & 0.7465 & 292.14 & 46.66  \\ \cline{2-8} 
                           & Label-only                 & Brep-MI & 38.50\%                     & 61.25\%                     & 0.9700 & 356.83 & 93.05  \\
\midrule
\multirow{4}{*}{\rotatebox{90}{\textbf{Facescrub}}} & White-box                  & PPA     & \multicolumn{1}{c}{\textbf{98.32\%}} & \multicolumn{1}{c}{\textbf{99.84\%}} & 0.6735 & 107.35 & 45.73  \\ \cline{2-8} 
                           & \multirow{2}{*}{Black-box} & RLB-MI  & \multicolumn{1}{c}{33.28\%} & \multicolumn{1}{c}{64.52\%} & 1.1097 & 135.16 & 111.06 \\
                           &                            & \textbf{CG-MI(Ours)}   & \multicolumn{1}{c}{\textbf{90.92\%}} & \multicolumn{1}{c}{\textbf{99.34\%}} & 0.7570 & 111.75 & 62.24  \\ \cline{2-8} 
                           & Label-only                 & Brep-MI & \multicolumn{1}{c}{51.33\%} & \multicolumn{1}{c}{73.82\%} & 1.0664 & 132.59 & 102.94 \\
\bottomrule
\end{tabular}%
}
\caption{Different MIA methods were applied to attack a Resnet18 model trained on CelebA and Facescrub datasets with $P(X_{prior})$ = FFHQ. In the black-box scenario, CG-MI significantly outperforms existing methods.}
\label{comparisiontable}
\end{table}

\Cref{comparisiontable} presents the evaluation results of CG-MI and baseline methods in attacking Resnet18 trained on Celeba and Facescrub datasets. The target model Resnet18 achieves test accuracies of 86.38\% and 94.22\% on Celeba and Facescrub, respectively, while the evaluation model InceptionV3 achieves test accuracies of 93.28\% and 96.20\% on the same datasets. Based on the data in \Cref{comparisiontable}, CG-MI outperforms other black-box methods in addressing distribution shift issues in black-box attack scenarios. It generates more transferability synthetic images, resulting in higher attack success rates, lower feature distances, and FID values.

The qualitative evaluation results shown in \Cref{comparisonresult} demonstrate that compared to previous black-box MIA methods, CG-MI is capable of generating more realistic images in different data distribution scenarios. It overcomes the limitations of distribution shift and produces synthetic images of comparable quality to state-of-the-art white-box methods.

\textbf{Performance Evaluation on various Models and Datasets.}
We also evaluated the performance of CG-MI on deeper network architectures and datasets from different categories. Specifically, we trained Resnet152 and Densenet169 models on the CelebA, Facescrub, and Stanford Dogs datasets. Additionally, we trained Resnet18 models on the Facescrub and CelebA datasets, with the recognition accuracy on the respective test datasets indicated in parentheses. 
\begin{figure}[t]
  \centering
   \includegraphics[width=1.0\linewidth]{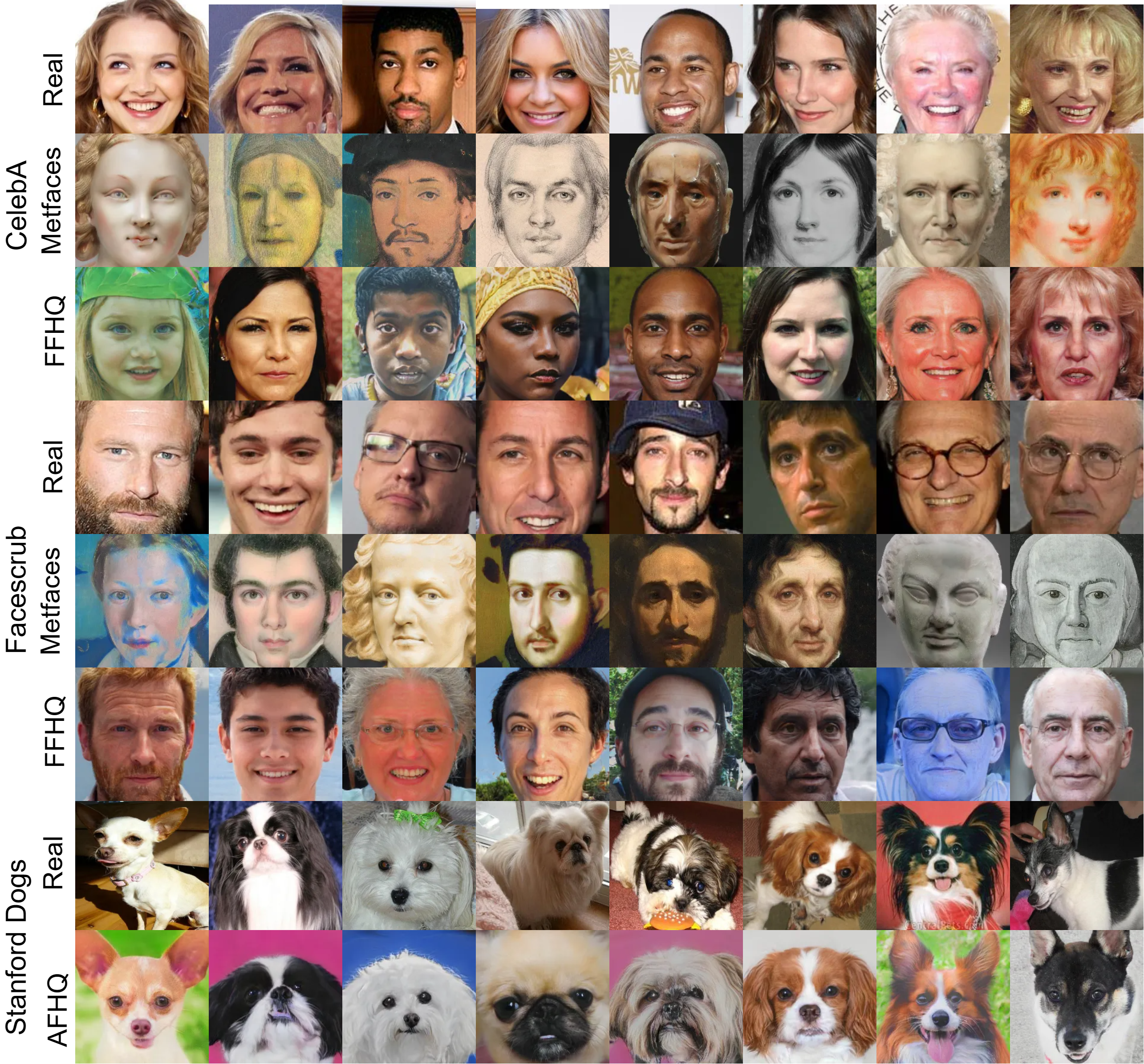}

   \caption{We have visualized the CG-MI attack results on the Densenet169 network architecture for the CelebA, Facescrub, and Stanford Dogs datasets.}
   \label{extendvisual}
\end{figure}

\begin{table}
\renewcommand{\arraystretch}{1.6}
\centering
\resizebox{\columnwidth}{!}{%
\begin{tabular}{@{}lccc@{}}

\toprule
\multicolumn{1}{l}{$P(X_{\textbf{prior}}) \rightarrow P(X_{\textbf{target}})$} &
\multicolumn{1}{l}{\textbf{Target Model}} &
\multicolumn{1}{c}{(\textbf{↑acc@1},\textbf{↑acc@5})} &
\multicolumn{1}{c}{(↓$\delta_{\textbf{face}}$, ↓$\delta_{\textbf{eval}}$, ↓$\textbf{FID}$)} \\

\bottomrule
\multicolumn{1}{l}{FFHQ $\rightarrow$ Celeba} & \multicolumn{1}{l}{Resnet152 (86.78\%)} & (\textbf{67.42\%},86.16\%) & (\textbf{0.7773, 319.42, 46.04}) \\
\multicolumn{1}{l}{FFHQ $\rightarrow$ Celeba} & \multicolumn{1}{l}{Densenet169 (85.39\%)} & (65.28\%, \textbf{87.72\%}) & (0.7831, 321.16, 47.91) \\
\multicolumn{1}{l}{Metfaces $\rightarrow$ Celeba} & \multicolumn{1}{l}{Resnet18 (86.38\%)} & (24.42\%, 49.26\%) & (1.2089, 420.75, 111.40) \\
\hline
\multicolumn{1}{l}{FFHQ $\rightarrow$ Facescrub} & \multicolumn{1}{l}{Resnet152 (93.74\%)} & (85.02\%, 97.94\%) & (0.7998, 122.53, 64.17) \\
\multicolumn{1}{l}{FFHQ $\rightarrow$ Facescrub} & \multicolumn{1}{l}{Densenet169 (95.49\%)} & (\textbf{90.78\%}, \textbf{97.95\%}) & \textbf{(0.7608, 115.02, 63.05)} \\
\multicolumn{1}{l}{Metfaces $\rightarrow$ Facescrub} & \multicolumn{1}{l}{Resnet18 (94.22\%)} & (59.16\%, 88.40\%) & (1.0286, 133.42, 101.02) \\
\hline
\multicolumn{1}{l}{AFHQ.dogs $\rightarrow$ Stan.dogs} & \multicolumn{1}{l}{Resnet152 (71.23\%)} & (\textbf{ 84.80\%}, \textbf{99.04\%}) & (-, \textbf{60.50, 58.04}) \\
\multicolumn{1}{l}{AFHQ.dogs $\rightarrow$ Stan.dogs} & \multicolumn{1}{l}{Densenet169 (74.39\%)} & (83.06\%, 98.02\%) & (-, 63.51, 59.03) \\
\hline
\end{tabular}
}
\caption{The attack results of CG-MI on different network architectures and datasets are evaluated. We employ the prior distribution $P(X_{prior})$ = FFHQ to attack the target distribution $P(X_{target})$ = Celeba and Facescrub, $P(X_{prior})$ = AFHQ.dogs to attack $P(X_{target})$ = St.dogs and $P(X_{prior})$ = Metfaces to attack $P(X_{target})$ = Celeba and Facescrub.}
\label{extendtable}
\end{table}

\begin{table}[t]
\centering
\resizebox{\columnwidth}{!}{%
\begin{tabular}{@{}lccccc@{}}
\toprule
Optimizer & ↑acc@1  & ↑acc@5  & ↓$\delta_{face}$ & ↓$\delta_{eval}$ & ↓$\text{FID}$ \\ 
\midrule
BO        & 54.65\%       & 85.26\%       & 0.9511                & 135.06                & 72.67      \\
CMA-ES    & \textbf{90.92\%} & \textbf{99.34\%} & \textbf{0.7570}           & \textbf{111.75 }          & \textbf{62.24}  \\ \hline
\end{tabular}
}
\caption{Attack performance with various gradient-free optimizer}
\label{Gradident-freeOptimizer}
\end{table}
The objective of this experiment was to explore the potential for our method to generalize to other model architectures or deeper network structures. It also aimed to evaluate the performance of CG-MI in scenarios involving significant data distribution shifts. From the data presented in \Cref{extendtable}, it is evident that as the target model becomes structurally deeper, CG-MI encounters increased difficulty when attacking CelebA and Facescrub. Consequently, this results in a slight reduction in attack effectiveness. In settings with significant data distribution shifts, such as using Metfaces to attack Celeba and Facescrub, CG-MI still maintains a certain level of attack effectiveness. In \Cref{extendvisual}, the visual experimental results illustrate that CG-MI continues to produce meaningful attack results, even when dealing with deeper network architectures, different classification tasks, and scenarios involving significant data distribution shifts. This further reinforces the robustness and versatility of our proposed method across various model architectures and datasets.

\textbf{Experiments with Various Gradient-free Optimizer.}
We compared different gradient-free optimization algorithms, namely CMA-ES\cite{hansen2016cma} and BO\cite{Eriksson_Pearce_Gardner_Turner_Poloczek_2019}, under the setting of $P(X_{prior})=FFHQ$ and $P(X_{target})=Facescrub$, with a target model architecture of Resnet18. Bayesian Optimization (BO) is a global optimization algorithm that utilizes probabilistic models to efficiently search for the optimal solution of an expensive black-box function. 

\subsection{Ablation Study}
In the ablation study, we considered two probability distributions: $P(X_{target})$, which represents the distribution of images from the CelebA dataset, and $P(X_{prior})$, which represents the distribution of images from the FFHQ dataset. We first compared three different loss functions for the model inversion attack: Poincaré loss\cite{struppek2022ppa}, max-margin loss\cite{yuan2023pseudo}, and cross-entropy loss\cite{chen2021knowledge}. Poincaré loss achieved the best performance in terms of generating effective synthesis images. 

The experiments involved also replacing the StyleGAN2 architecture with the DCGAN architecture combine our proposed attack CG-MI. Ablation study results indicated that CG-MI is compatible with other GAN architectures. Our findings demonstrate that, even when replacing StyleGAN2 with the DCGAN, CG-MI still yields favorable evaluation results.
We also conducted ablation experiments by employing the objective function proposed in \cite{struppek2022ppa} in combination with the gradient-free optimization algorithm CMA-ES. From the experimental results(\textbf{No Mapping}), we observed that in the black-box scenario, where gradient information is unavailable, directly using the objective function from PPA did not yield favorable attack results. This underscores the importance of leveraging the mapping network as an integral part of the optimization process. By using our newly proposed objective function, we significantly improved our attack capabilities in the black-box scenario.

We also investigated the influence of transformation-based selection techniques\cite{struppek2022ppa} on the attack outcomes (\textbf{No Trans. Selection}). Transformation-based selection involves applying certain transformations to the synthesis images after the optimization process. By employing selection transformations, we can enhance the stability of attack outcomes and, to some extent, increase the success rate of attacks. The results of our ablation experiments highlighted the significance of using our proposed objective function for generating synthesis images. 

\begin{table}[t]
\resizebox{\columnwidth}{!}{%
\begin{tabular}{@{}lccccc@{}}
\toprule
                     & \textbf{↑acc@1}  & \textbf{↑acc@5}  & ↓$\delta_{\textbf{face}}$ & ↓$\delta_{\textbf{eval}}$ & ↓$\textbf{FID}$ \\ 
\bottomrule                     
\textbf{Poincare Loss}        & \textbf{77.86\%} & \textbf{94.16\%} & \textbf{0.7465}           & \textbf{292.14}           & \textbf{46.66}  \\
\textbf{Cross-Entropy Loss}   & 62.77\% & 85.67\% & 0.8550           & 329.71           & 51.81  \\
\textbf{Max-Margin Loss}      & 71.50\% & 91.25\% & 0.7827           & 331.48           & 49.02  \\
\textbf{DCGAN}  &32.80\% &66.50\% & 0.9526  
& 322.88 & 91.32 \\
\textbf{No Mapping} & \textbf{00.03\%} & \textbf{00.09\%} &1.4894 &435.82 &201.02 \\
\textbf{No Trans. Selection}   & 72.79\% & 89.31\% & 0.7688           & 300.11           & 47.28  \\ \hline
\end{tabular}
}
\caption{Ablation study performed on a Resnet18 trained on CelebA using the FFHQ StyleGAN2 as image prior.}
\label{Ablation Study Table}
\end{table}

\section{Discussion, Limitations and Conclusion}
Our paper proposes a novel black-box attack method called CG-MI. Unlike existing black-box methods, we focus on considering more realistic scenarios without making assumptions about the data distribution of the target model. Our approach only requires knowledge about the model’s classification task and leverages pre-trained publicly available images’ prior knowledge to attack various target models and generate high-resolution synthetic images. Furthermore, we introduce the concept of synthetic images transferability and investigate its impact on in MIAs. By designing a novel objective function and combining gradient-free optimization methods, we achieve MIAs in black-box scenarios and enhance the transferability of the synthesized images. Experimental results demonstrate that CG-MI outperforms existing black-box MIAs in more realistic scenarios, achieving state-of-the-art attack performance.

However, the current black-box MIAs, including our work, still have some limitations. While black-box methods do not require full access to the target model, frequent queries to the target model may hinder the progress of the attack in real-world settings. Therefore, exploring how to control the query count and ensure attack success rate is a worthwhile research direction. 

It is worth noting that our research may have negative implications. However, the purpose of revealing vulnerabilities in existing systems is to promote the development of better defense mechanisms. Our work aims to call for attention from the academic and technical community to research on machine learning privacy protection. We believe that the positive impact of these efforts will outweigh the potential negative risks.
\section*{Acknowledgements}
\clearpage
{
    \small
    \bibliographystyle{ieeenat_fullname}
    \bibliography{main}

\begin{thebibliography}{41}
\providecommand{\natexlab}[1]{#1}
\providecommand{\url}[1]{\texttt{#1}}
\expandafter\ifx\csname urlstyle\endcsname\relax
  \providecommand{\doi}[1]{doi: #1}\else
  \providecommand{\doi}{doi: \begingroup \urlstyle{rm}\Url}\fi

\bibitem[An et~al.(2022)An, Tao, Xu, Liu, Shen, Yao, Xu, and Zhang]{mirror}
Shengwei An, Guanhong Tao, Qiuling Xu, Yingqi Liu, Guangyu Shen, Yuan Yao, Jingwei Xu, and Xiangyu Zhang.
\newblock Mirror: Model inversion for deep learning network with high fidelity.
\newblock In \emph{Proceedings of the 29th Network and Distributed System Security Symposium}, 2022.

\bibitem[Aïvodji et~al.(2019)Aïvodji, Gambs, and Ther]{black-box}
Ulrich Aïvodji, Sébastien Gambs, and Timon Ther.
\newblock Gamin: An adversarial approach to black-box model inversion.
\newblock \emph{Cornell University - arXiv,Cornell University - arXiv}, 2019.

\bibitem[Brock et~al.(2018)Brock, Donahue, and Simonyan]{biggan}
AndrewS. Brock, Jeff Donahue, and Karen Simonyan.
\newblock Large scale gan training for high fidelity natural image synthesis.
\newblock \emph{International Conference on Learning Representations}, 2018.

\bibitem[Chen et~al.(2021)Chen, Kahla, Jia, and Qi]{chen2021knowledge}
Si Chen, Mostafa Kahla, Ruoxi Jia, and Guo-Jun Qi.
\newblock Knowledge-enriched distributional model inversion attacks.
\newblock In \emph{Proceedings of the IEEE/CVF international conference on computer vision (CVPR)}, pages 16178--16187, 2021.

\bibitem[Choi et~al.(2020)Choi, Uh, Yoo, and Ha]{afhq}
Yunjey Choi, Youngjung Uh, Jaejun Yoo, and Jung-Woo Ha.
\newblock Stargan v2: Diverse image synthesis for multiple domains.
\newblock In \emph{2020 IEEE/CVF Conference on Computer Vision and Pattern Recognition (CVPR)}, 2020.

\bibitem[Das and Suganthan(2011)]{DE}
Swagatam Das and Ponnuthurai~Nagaratnam Suganthan.
\newblock Differential evolution: A survey of the state-of-the-art.
\newblock \emph{IEEE Transactions on Evolutionary Computation}, page 4–31, 2011.

\bibitem[Deng et~al.(2009)Deng, Dong, Socher, Li, Li, and Fei-Fei]{imagenet}
Jia Deng, Wei Dong, Richard Socher, Li-Jia Li, Kai Li, and Li Fei-Fei.
\newblock Imagenet: A large-scale hierarchical image database.
\newblock In \emph{2009 IEEE Conference on Computer Vision and Pattern Recognition (CVPR)}, 2009.

\bibitem[Eriksson et~al.(2019)Eriksson, Pearce, Gardner, Turner, and Poloczek]{Eriksson_Pearce_Gardner_Turner_Poloczek_2019}
David Eriksson, Michael Pearce, JacobR. Gardner, Ryan Turner, and Matthias Poloczek.
\newblock Scalable global optimization via local bayesian optimization.
\newblock \emph{Neural Information Processing Systems(NeurIPS)}, 2019.

\bibitem[Fredrikson et~al.(2014)Fredrikson, Lantz, Jha, Lin, Page, and Ristenpart]{firstpaper2014}
Matthew Fredrikson, Eric Lantz, Somesh Jha, Simon Lin, David Page, and Thomas Ristenpart.
\newblock Privacy in pharmacogenetics: An end-to-end case study of personalized warfarin dosing.
\newblock In \emph{23rd $\{$USENIX$\}$ Security Symposium ($\$USENIX$\$ Security 14)}, pages 17--32, 2014.

\bibitem[Fredrikson et~al.(2015)Fredrikson, Jha, and Ristenpart]{first-paper}
Matt Fredrikson, Somesh Jha, and Thomas Ristenpart.
\newblock Model inversion attacks that exploit confidence information and basic countermeasures.
\newblock In \emph{Proceedings of the 22nd ACM SIGSAC conference on computer and communications security}, pages 1322--1333, 2015.

\bibitem[Goodfellow et~al.(2017)Goodfellow, Pouget-Abadie, Mirza, Xu, Warde-Farley, Ozair, Courville, and Bengio]{Gan}
Ian Goodfellow, Jean Pouget-Abadie, Mehdi Mirza, Bing Xu, David Warde-Farley, Sherjil Ozair, Aaron Courville, and Yoshua Bengio.
\newblock Generative adversarial nets.
\newblock \emph{Journal of Japan Society for Fuzzy Theory and Intelligent Informatics}, page 177–177, 2017.

\bibitem[Han et~al.(2023)Han, Choi, Lee, and Kim]{han2023reinforcement}
Gyojin Han, Jaehyun Choi, Haeil Lee, and Junmo Kim.
\newblock Reinforcement learning-based black-box model inversion attacks.
\newblock In \emph{Proceedings of the IEEE/CVF Conference on Computer Vision and Pattern Recognition (CVPR)}, pages 20504--20513, 2023.

\bibitem[Hansen(2016)]{hansen2016cma}
Nikolaus Hansen.
\newblock The cma evolution strategy: A tutorial.
\newblock \emph{Towards a new evolutionary computation}, page 75–102, 2016.

\bibitem[Hansen et~al.(2019)Hansen, Akimoto, and Baudis]{pycma}
Nikolaus Hansen, Youhei Akimoto, and Petr Baudis.
\newblock {CMA-ES/pycma} on {G}ithub.
\newblock Zenodo, DOI:10.5281/zenodo.2559634, 2019.

\bibitem[He et~al.(2016)He, Zhang, Ren, and Sun]{resnet}
Kaiming He, Xiangyu Zhang, Shaoqing Ren, and Jian Sun.
\newblock Deep residual learning for image recognition.
\newblock In \emph{2016 IEEE Conference on Computer Vision and Pattern Recognition (CVPR)}, 2016.

\bibitem[Heusel et~al.(2017)Heusel, Ramsauer, Unterthiner, Nessler, and Hochreiter]{fid}
Martin Heusel, Hubert Ramsauer, Thomas Unterthiner, Bernhard Nessler, and Sepp Hochreiter.
\newblock Gans trained by a two time-scale update rule converge to a local nash equilibrium.
\newblock \emph{Neural Information Processing Systems (NeurIPS)}, 2017.

\bibitem[Huang et~al.(2017)Huang, Liu, Van Der~Maaten, and Weinberger]{densenet}
Gao Huang, Zhuang Liu, Laurens Van Der~Maaten, and Kilian~Q. Weinberger.
\newblock Densely connected convolutional networks.
\newblock In \emph{2017 IEEE Conference on Computer Vision and Pattern Recognition (CVPR)}, 2017.

\bibitem[Kahla et~al.(2022)Kahla, Chen, Just, and Jia]{labelonly}
Mostafa Kahla, Si Chen, Hoang~Anh Just, and Ruoxi Jia.
\newblock Label-only model inversion attacks via boundary repulsion.
\newblock In \emph{Proceedings of the IEEE/CVF Conference on Computer Vision and Pattern Recognition (CVPR)}, pages 15045--15053, 2022.

\bibitem[Karras et~al.(2019)Karras, Laine, and Aila]{FFHQ}
Tero Karras, Samuli Laine, and Timo Aila.
\newblock A style-based generator architecture for generative adversarial networks.
\newblock In \emph{2019 IEEE/CVF Conference on Computer Vision and Pattern Recognition (CVPR)}, 2019.

\bibitem[Karras et~al.(2020{\natexlab{a}})Karras, Aittala, Hellsten, Laine, Lehtinen, and Aila]{metafaces}
Tero Karras, Miika Aittala, Janne Hellsten, Samuli Laine, Jaakko Lehtinen, and Timo Aila.
\newblock Training generative adversarial networks with limited data.
\newblock \emph{Neural Information Processing Systems,Neural Information Processing Systems}, 2020{\natexlab{a}}.

\bibitem[Karras et~al.(2020{\natexlab{b}})Karras, Aittala, Hellsten, Laine, Lehtinen, and Aila]{stylegan2-ada}
Tero Karras, Miika Aittala, Janne Hellsten, Samuli Laine, Jaakko Lehtinen, and Timo Aila.
\newblock Training generative adversarial networks with limited data.
\newblock \emph{Cornell University - arXiv}, 2020{\natexlab{b}}.

\bibitem[Karras et~al.(2020{\natexlab{c}})Karras, Laine, Aittala, Hellsten, Lehtinen, and Aila]{stylegan2}
Tero Karras, Samuli Laine, Miika Aittala, Janne Hellsten, Jaakko Lehtinen, and Timo Aila.
\newblock Analyzing and improving the image quality of stylegan.
\newblock In \emph{2020 IEEE/CVF Conference on Computer Vision and Pattern Recognition (CVPR)}, 2020{\natexlab{c}}.

\bibitem[Khosla et~al.(2011)Khosla, Jayadevaprakash, Yao, and Fei-Fei]{stanforddogs}
Aditya Khosla, N Jayadevaprakash, Bangpeng Yao, and Li Fei-Fei.
\newblock Novel dataset for fine-grained image categorization.
\newblock In \emph{Conference on Computer Vision and Pattern Recognition (CVPR) Workshop}, 2011.

\bibitem[Liu et~al.(2020)Liu, Xie, Wang, Zou, Xiong, Ying, and Vasilakos]{liu2020privacy}
Ximeng Liu, Lehui Xie, Yaopeng Wang, Jian Zou, Jinbo Xiong, Zuobin Ying, and Athanasios~V Vasilakos.
\newblock Privacy and security issues in deep learning: A survey.
\newblock \emph{IEEE Access}, 9:\penalty0 4566--4593, 2020.

\bibitem[Liu et~al.(2015)Liu, Luo, Wang, and Tang]{celeba}
Ziwei Liu, Ping Luo, Xiaogang Wang, and Xiaoou Tang.
\newblock Deep learning face attributes in the wild.
\newblock In \emph{2015 IEEE International Conference on Computer Vision (ICCV)}, 2015.

\bibitem[Ng and Winkler(2014)]{facescrub}
Hong-Wei Ng and Stefan Winkler.
\newblock A data-driven approach to cleaning large face datasets.
\newblock In \emph{2014 IEEE International Conference on Image Processing (ICIP)}, 2014.

\bibitem[Nguyen et~al.(2023)Nguyen, Chandrasegaran, Abdollahzadeh, and Cheung]{rethink}
Ngoc-Bao Nguyen, Keshigeyan Chandrasegaran, Milad Abdollahzadeh, and Ngai-Man Cheung.
\newblock Re-thinking model inversion attacks against deep neural networks.
\newblock In \emph{Proceedings of the IEEE/CVF Conference on Computer Vision and Pattern Recognition (CVPR)}, pages 16384--16393, 2023.

\bibitem[Radford et~al.(2016)Radford, Metz, and Chintala]{DCGAN}
Alec Radford, Luke Metz, and Soumith Chintala.
\newblock Unsupervised representation learning with deep convolutional generative adversarial networks.
\newblock \emph{International Conference on Learning Representations (ICLR)}, 2016.

\bibitem[Razzhigaev et~al.(2020)Razzhigaev, Kireev, Kaziakhmedov, Tursynbek, and Petiushko]{black-box-}
Anton Razzhigaev, Klim Kireev, Edgar Kaziakhmedov, Nurislam Tursynbek, and Aleksandr Petiushko.
\newblock Black-box face recovery from identity features.
\newblock In \emph{Computer Vision--ECCV 2020 Workshops: Glasgow, UK, August 23--28, 2020, Proceedings, Part V 16}, pages 462--475. Springer, 2020.

\bibitem[Rigaki and García(2020{\natexlab{a}})]{PrivacySurvey}
Maria Rigaki and Sebastián García.
\newblock A survey of privacy attacks in machine learning.
\newblock \emph{Cornell University - arXiv}, 2020{\natexlab{a}}.

\bibitem[Rigaki and García(2020{\natexlab{b}})]{rigaki2020survey}
Maria Rigaki and Sebastián García.
\newblock A survey of privacy attacks in machine learning.
\newblock \emph{Cornell University - arXiv,Cornell University - arXiv}, 2020{\natexlab{b}}.

\bibitem[Schroff et~al.(2015)Schroff, Kalenichenko, and Philbin]{facenet}
Florian Schroff, Dmitry Kalenichenko, and James Philbin.
\newblock Facenet: A unified embedding for face recognition and clustering.
\newblock In \emph{2015 IEEE Conference on Computer Vision and Pattern Recognition (CVPR)}, 2015.

\bibitem[Song and Namiot(2022)]{song2022survey}
Junzhe Song and Dmitry Namiot.
\newblock A survey of the implementations of model inversion attacks.
\newblock In \emph{International Conference on Distributed Computer and Communication Networks}, pages 3--16. Springer, 2022.

\bibitem[Struppek et~al.(2022)Struppek, Hintersdorf, Correia, Adler, and Kersting]{struppek2022ppa}
Lukas Struppek, Dominik Hintersdorf, Antonio De~Almeida Correia, Antonia Adler, and Kristian Kersting.
\newblock Plug \& play attacks: Towards robust and flexible model inversion attacks.
\newblock In \emph{Proceedings of the 39th International Conference on Machine Learning (ICML)}, pages 20522--20545. PMLR, 2022.

\bibitem[Szegedy et~al.(2016)Szegedy, Vanhoucke, Ioffe, Shlens, and Wojna]{inceptionv3}
Christian Szegedy, Vincent Vanhoucke, Sergey Ioffe, Jon Shlens, and Zbigniew Wojna.
\newblock Rethinking the inception architecture for computer vision.
\newblock In \emph{2016 IEEE Conference on Computer Vision and Pattern Recognition (CVPR)}, 2016.

\bibitem[Wang et~al.(2021)Wang, Fu, Li, Khisti, Zemel, and Makhzani]{vmi}
Kuan-Chieh Wang, Yan Fu, Ke Li, Ashish Khisti, Richard Zemel, and Alireza Makhzani.
\newblock Variational model inversion attacks.
\newblock \emph{Advances in Neural Information Processing Systems}, 34:\penalty0 9706--9719, 2021.

\bibitem[Wang et~al.(2018)Wang, Liu, Zhu, Tao, Kautz, and Catanzaro]{cgan}
Ting-Chun Wang, Ming-Yu Liu, Jun-Yan Zhu, Andrew Tao, Jan Kautz, and Bryan Catanzaro.
\newblock High-resolution image synthesis and semantic manipulation with conditional gans.
\newblock In \emph{Proceedings of the IEEE conference on computer vision and pattern recognition (CVPR)}, pages 8798--8807, 2018.

\bibitem[Wernke et~al.(2014)Wernke, Skvortsov, D{\"u}rr, and Rothermel]{wernke2014classification}
Marius Wernke, Pavel Skvortsov, Frank D{\"u}rr, and Kurt Rothermel.
\newblock A classification of location privacy attacks and approaches.
\newblock \emph{Personal and ubiquitous computing}, 18:\penalty0 163--175, 2014.

\bibitem[Yang et~al.(2019)Yang, Chang, and Liang]{black-boxlbmi}
Ziqi Yang, Ee-Chien Chang, and Zhenkai Liang.
\newblock Neural network inversion in adversarial setting via background knowledge alignment.
\newblock \emph{In Proceedings of the 2019 ACM SIGSAC Conference on Computer and Communications Security, CCS '19}, page 225–240, 2019.

\bibitem[Yuan et~al.(2023)Yuan, Chen, Zhang, Zhang, Yu, and Zhang]{yuan2023pseudo}
Xiaojian Yuan, Kejiang Chen, Jie Zhang, Weiming Zhang, Nenghai Yu, and Yang Zhang.
\newblock Pseudo label-guided model inversion attack via conditional generative adversarial network.
\newblock \emph{AAAI Conference on Artificial Intelligence(AAAI)}, 2023.

\bibitem[Zhang et~al.(2020)Zhang, Jia, Pei, Wang, Li, and Song]{zhang2020secret}
Yuheng Zhang, Ruoxi Jia, Hengzhi Pei, Wenxiao Wang, Bo Li, and Dawn Song.
\newblock The secret revealer: Generative model-inversion attacks against deep neural networks.
\newblock In \emph{Proceedings of the IEEE/CVF conference on computer vision and pattern recognition (CVPR)}, pages 253--261, 2020.

\end{thebibliography}
}
\clearpage
\setcounter{page}{1}
\maketitlesupplementary

\begin{appendices}
\begin{figure*}[t]
  \centering
   \includegraphics[width=1.0\linewidth]{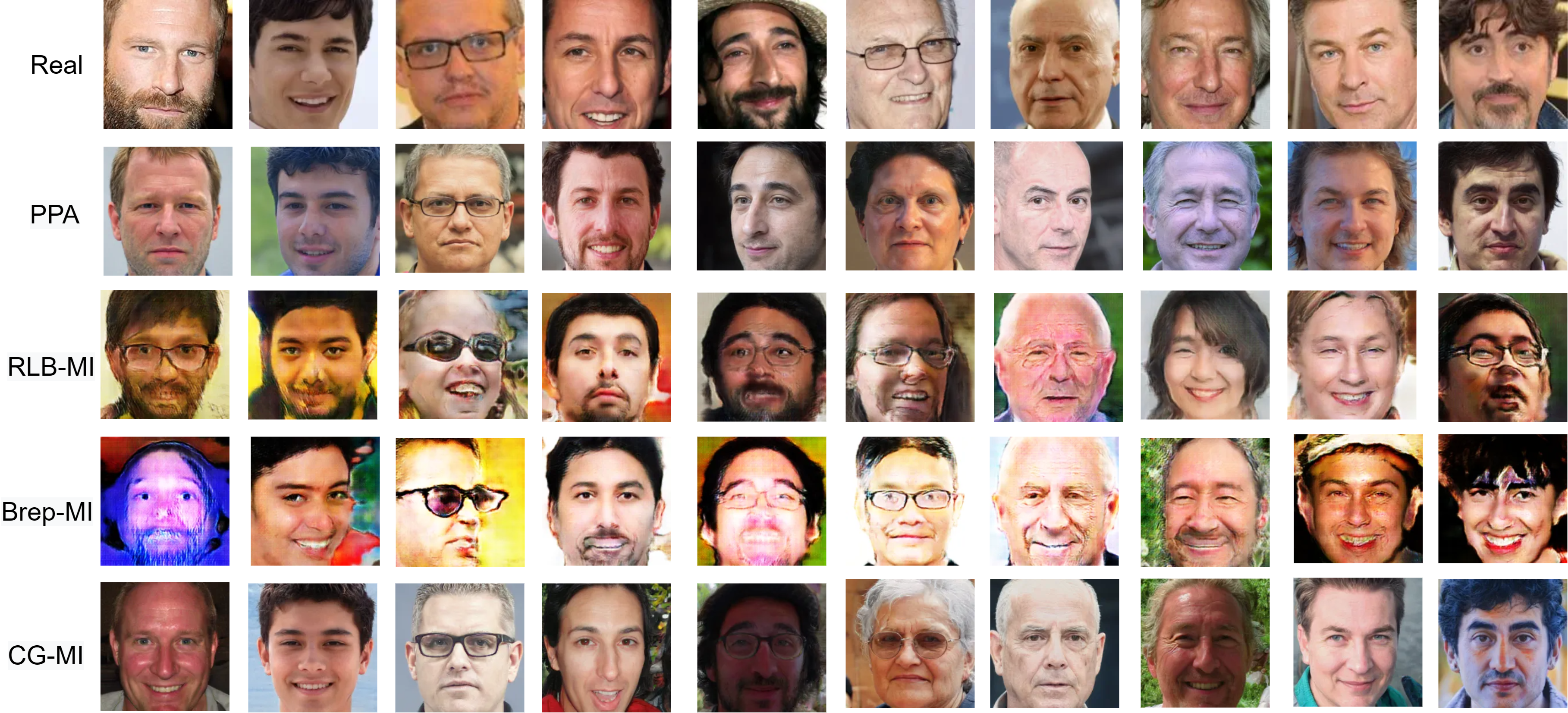}
   \caption{We present a visual comparison of the attack results for different methods in the scenario where the $P(X_{prior})=FFHQ$, $P(X_{target})=Facescrub$ and the target model architecture is Resnet18. The first row shows ground truth images of target class. The second row represents PPA\cite{struppek2022ppa}, the third row represents RLB-MI\cite{han2023reinforcement}, and the fourth row represents Brep-MI\cite{labelonly}. The last row introduces our proposed method, CG-MI.} 
   \label{comparisonresultfacescrub}
\end{figure*}
\leftline{\textbf{Appendix}}

\section{Confidence Matching Loss } \label{loss}
We explore three key loss functions: cross-entropy loss\cite{zhang2020secret,chen2021knowledge}, max-margin loss\cite{yuan2023pseudo} and Poincar\'{e} loss\cite{struppek2022ppa}. Inspired by \cite{rethink}, we represent the model’s unnormalized scores by taking the product of the weights of the last layer and the penultimate layer activations . The cross-entropy loss aims to minimize the negative log-likelihood of the identity under the model parameters. The formulation is as follows:
\begin{equation}
    \mathcal{L}_{ce} = -\log{\left(\frac{\exp{\left(p^\intercal w_c\right)}}{\exp{\left(p^\intercal w_c\right)}+\sum_{j=1,j\neq c}\exp{\left(p^\intercal w_j\right)}}\right)}
\end{equation}
Here, $p$ denotes the activation of the penultimate layer for sample $x$, and $w_c$ represents the weight of the last layer for the $c$-th class in the target model $M$.

The maximum margin loss not only encourages maximizing the confidence score for the target class $c$ but also emphasizes the separability of this class from others. We reformulate the maximum margin loss as follows:
\begin{equation}
    \mathcal{L}_{mm} = -\left(p^\intercal w_c\right) + \max_{j\neq c}\left(p^\intercal w_j\right)
\end{equation}
Here, $p$ represents the activation of the penultimate layer for sample $x$. The weight of the last layer for the $c$-th class in the target model $M$ is denoted as $w_c$. The term $p^\intercal w_c$ represents the unnormalized logit value for the $c$-th class.

Poincar\'{e} loss is a hyperbolic space embedding loss function, which can be rewritten as:
\begin{equation}
\begin{aligned}
    \mathcal{L}_{p}
    = \text{arcosh}\left(1 + 2\frac{\left|\left|\frac{p^Tw_c} {\left|\left|p^Tw_c\right|\right|_1}-y_c\right|\right|_2^2}{\left(1-\left|\left|\frac{p^Tw_c}{\left|\left|p^Tw_c\right|\right|_1}\right|\right|_2^2\right)\left(1-\left|\left|y_c\right|\right|_2^2\right)}\right)
\end{aligned}
\end{equation}
Poincar\'{e} loss measures the distance between two vectors $u$ and $v$ in the hyperbolic space. $||\cdot||_2$ represents the Euclidean norm, satisfying $||u||_2<1$ and $||v||_2<1$. Here, $u=\frac{p^Tw_c}{||p^Tw_c||_1}$, $||\cdot||_1$ denotes the absolute value norm, $v$ is the one-hot encoded vector for class $c$, denoted as $y_c$, and we replace 1 with 0.9999. Poincaré loss belongs to the hyperbolic distance learning paradigm, which enables measuring and comparing distances between vectors in a larger embedding space.

\section{Experimental Supplement}

\subsection{Datasets}\label{detail_dataset}
CelebA is a dataset of celebrity face attributes that contains 202,599 face pictures of 10,177 celebrity identities. 
For the training data of the target models (Resnet18, Resnet152, and Densenet169), we selected 1000 identities with the most number of samples, resulting in a total of 30,038 images. The FaceScrub dataset provides cropped face images of 530 identities. However, on the dataset’s official website, instead of actual images, they provide download links for the dataset. All identities are used as target dataset. For Stanford.dogs,  this dataset is built on top of ImageNet, which is intended for non-commercial research purposes only and provides 20,580 images of 120 dog breeds. For all datasets, the input images for the target model are resized to 224x224, while the input images for the evaluation model(Inception-V3) are resized to 299x299. 
The CelebA\cite{celeba} dataset comprises 202,599 face pictures of 10,177 celebrity identities. In the case of training the target models (Resnet18, Resnet152, and Densenet169), we specifically selected 1,000 identities with the highest number of samples, resulting in a total of 30,038 images. Please note that the FaceScrub\cite{facescrub} dataset provides cropped face images of 530 identities, but the dataset’s official website only offers download links instead of actual images. All identities from FaceScrub were utilized as a target dataset for our research. Regarding the Stanford.dogs\cite{stanforddogs} dataset, it is constructed on the foundation of ImageNet\cite{imagenet}, which is exclusively intended for non-commercial research purposes. The ImageNet dataset provides 20,580 images encompassing 120 dog breeds. In our experiments, the input images for the target models were uniformly resized to dimensions of 224x224 pixels, ensuring consistency across all datasets. However, it is important to note that for the evaluation model (Inception-V3), the input images were resized to a different size of 299x299 pixels.

\textbf{Attack Implementation.}\label{attackimplementation}
All models were trained using the Adam optimizer with a learning rate of 0.001 and $\beta$ values of $[0.9, 0.999]$ for a total of 100 epochs with a batch size of 128. The training data was normalized with mean $\mu$ and standard deviation $\sigma$ set to 0.5. The input images of the target model were resized to 224$\times$224, and the evaluation model InceptionV3 was resized to 299$\times$299. Additionally, data augmentation techniques were applied, including 50\% horizontal flipping and adjustments in brightness and contrast within the range $[0.8, 1.2]$, saturation within $[0.9, 1.1]$, and hue within $[-0.1, 0.1]$.

During the attack process, the StyleGAN2 synthesis network was configured with a truncation parameter $\psi$ of 0.5 and a truncation cutoff value of 8. The CMA-ES algorithm, a gradient-free optimization method, was employed with 8 iterations, 300 rounds, and a population size of 25. The rotation transformation selection strategy involved 100 transformations. It included center cropping of images generated by the generative model, resizing them to 224, and applying random adjustments to cropping parameters within the ranges of size $[224, 224]$, scale $[0.5, 0.9]$, and ratio $[0.8, 1.2]$. To accelerate the multiple CMA algorithm optimizations, 8 parallel processes were utilized. In the extended experiments, we performed central cropping to 800 on the images generated by Metfaces for Celeba and Facescrub datasets, followed by resizing to 224. For the remaining different architectures of target models, we maintained consistent attack parameters.

In the comparative experiments to perform RLB-MI and Brep-MI, the latent vector size in the GAN was set to 100. For GAN training, the ADAM optimizer was used with a batch size of 64, a learning rate of 0.0002, and $\beta$ values of $[0.5, 0.999]$. The training was conducted for 280 epochs.

\subsection{Publicly Available Image Prior} 
We downloaded the code for StyleGAN2\cite{stylegan2-ada} from the official source at \href{https://github.com/NVlabs/stylegan2-ada-pytorch}{stylegan2-ada}. The AFHQ dataset consists of 16,130 high-resolution images with a resolution of 512x512 pixels. We obtained the pretrained model weights for AFHQ.dogs512\cite{afhq} by using the provided link from the official code of StyleGAN2-ADA: \href{https://nvlabs-fi-cdn.nvidia.com/stylegan2-ada-pytorch/pretrained/afhqdog.pkl}{AFHQ.dogs}. 
It should be noted that the FFHQ\cite{FFHQ} dataset comprises 70,000 high-quality face images. Compared to CelebA and FaceScrub, the image quality in FFHQ is significantly higher. Subsequently, we downloaded the pretrained model weights for FFHQ256 from \href{https://nvlabs-fi-cdn.nvidia.com/stylegan2-ada-pytorch/pretrained/transfer-learning-source-nets/ffhq-res256-mirror-paper256-noaug.pkl}{ffhq256} and for Metfaces from \href{https://nvlabs-fi-cdn.nvidia.com/stylegan2-ada-pytorch/pretrained/}{Metfaces}. During the attack process, for the StyleGAN2 model pretrained on FFHQ256, we performed central cropping to 200 and then resized the images to 224x224 before inputting them into the target model. For the generated images on AFHQ.dogs, we applied central cropping to 400 and then resized them to 224x224. In the comparative experiments, we followed the same cropping and resizing approach for other MIAs method.

\begin{algorithm}[t]
\SetAlgoNlRelativeSize{0}
\caption{CMA-ES Algorithm for MIA}
\label{weidaima}
\KwIn{Population size $\lambda$, Learning rate for covariance matrix update $c_1$, Learning rate for step-size adaptation $c_{\sigma}$, $d_{\sigma}$, Initial mean vector $\mathbf{m}$, Initial step size $\sigma$, Covariance matrix $\mathbf{C}$, Fitness based weight $w$, Target model $M$, Target class $c$, Synthesis network $G_{\text{synthesis}}$, Mapping network $G_{\text{mapping}}$, Maximum iterations $T$}
\KwOut{Optimized latent vector $z^*$}

\While{current step $t < T$}{
  Generate $\lambda$ latent vectors \{${z_0, ..., z_{\lambda-1}}$\} from multivariate normal distribution with mean $\mathbf{m}$ and covariance $\mathbf{C}$\;
  Evaluate fitness based on $\mathcal{L}(M(G_{\text{synthesis}}(G_{\text{mapping}}(z_i)), c))$, where $z_i \in \{z_0, ..., z_{\lambda-1}\}$\;
  Select the top $\mu$ latent vectors with the highest fitness\;
  Update mean vector:\\
  \hspace{1em} $\mathbf{m}_{\text{new}} = \mathbf{m} + (1/\mu) \sum_{i=1}^{\mu} w_i \cdot  \mathbf{z}_i$\;
  Update covariance matrix:\\
  \hspace{1em} $\mathbf{u}_i = \mathbf{C}^{-1} \cdot (\mathbf{z}_i - \mathbf{m})$\;
  \hspace{1em} $\mathbf{C}_{\text{new}} = (1 - c_1) \cdot \mathbf{C} + c_1 \sum_{i=1}^{\mu} w_i \cdot \mathbf{u}_i \cdot \mathbf{u}_i^T$\;
  Update step size:\\
  \hspace{1em} $\sigma_{\text{new}} = \sigma \cdot \exp\left(\frac{c_{\sigma}}{d_{\sigma}} \left(\frac{\|\mathbf{C}\|}{E[\|\mathbf{N}(0, \mathbf{I})\|]} - 1\right)\right)$\;
  Update current solution based on $m_{\text{new}}$ and $\sigma_{\text{new}}$\;
  current step $t++$\;
}
\end{algorithm}

\subsection{Additional Experimental Results.}

\textbf{Visual Comparison of Attack Results on Facescrub.}
We have also visualized the attack results using the FaceScrub dataset, comparing the performance of different methods. From the visualized results in \cref{comparisonresultfacescrub}, our approach demonstrates the ability to generate synthesis images that better reflect the distinctive features of the target model’s private training data.
\end{appendices}



\end{document}